%% file: main.tex
\documentclass[10pt,twocolumn,letterpaper]{article}

\usepackage{iccv}
\usepackage{times}
\usepackage{epsfig}
\usepackage{graphicx}
\usepackage{amsmath}
\usepackage{bbm}
\usepackage{amssymb}
\usepackage{siunitx}
\usepackage{cancel}
\usepackage{mycommands}
\usepackage{color}
\usepackage{pifont}
\newcommand{\cmark}{\ding{51}}%
\usepackage{tabularx}
\usepackage{xpatch}
\usepackage{enumitem}
\usepackage[accsupp]{axessibility}

\usepackage{floatrow}
\usepackage{paralist}

\usepackage[pagebackref=true,breaklinks=true,letterpaper=true,colorlinks,bookmarks=false]{hyperref}

\iccvfinalcopy %

\def\iccvPaperID{9433} %

\ificcvfinal\fi
\begin{document}
\urlstyle{same}

\input{sections/title.tex}

\input{sections/00_abstract.tex}

\input{sections/intro_new.tex}

\input{sections/02_rw.tex}
\input{sections/03_retrieval.tex}

\input{sections/04_reranking.tex}

\input{sections/05_experiments.tex}

\input{sections/06_conclusion.tex}

\appendix
\section*{Appendix}
\input{sections/S4.tex}

\input{sections/S5.tex}
\input{sections/S6.tex}
\input{sections/S7.tex}

{\small
\bibliographystyle{ieee_fullname}
\bibliography{main}
}

\end{document}

\title{Global Features are All You Need for Image Retrieval and Reranking\\Supplementary Material}

\maketitle
\thispagestyle{empty}

\section{Abalation Studay}
\subsection{GeM+}
\subsection{Scale-GeM}
\subsection{Regional-GeM}
\label{sec:reranking}

\section{Regional-GeM}
\section{Rerank top-800 and top-1600}

%% file: sections/title.tex
\title{Global Features are All You Need for Image Retrieval and Reranking}

\title{Global Features are All You Need for Image Retrieval and Reranking}

\author{Shihao Shao$^1$ \hspace{-0.4em}  \footnotemark[1] \hspace{-0.10em} \footnotemark[2]
\and
Kaifeng Chen$^2$
\and
Arjun Karpur$^2$
\and
Qinghua Cui$^1$
\and
André Araujo$^2$
\and
Bingyi Cao$^2$ \hspace{-0.4em}  \footnotemark[1] \hspace{-0.10em} \footnotemark[2] \\
\vspace{-4pt}
$^1$Peking University \hspace{20pt} $^2$Google Research
}

\maketitle
\makeatletter
\renewcommand\paragraph{\@startsection{paragraph}{4}{\z@}%
                                    {0.5ex \@plus1ex \@minus.2ex}%
                                    {-1em}%
                                    {\normalfont\normalsize\bfseries}}
\renewcommand{\thefootnote}{\fnsymbol{footnote}}
\footnotetext[1]{Both authors contributed equally to this paper.}
\footnotetext[2]{Co-corresponding authors. Emails:}
\footnotetext{shaoshihao@pku.edu.cn, bingyi@google.com}
\makeatother

%% file: sections/00_abstract.tex
\begin{abstract}

Image retrieval systems conventionally use a two-stage paradigm, leveraging global features for initial retrieval and local features for reranking. 
However, the scalability of this method is often limited due to the significant storage and computation cost incurred by local feature matching in the reranking stage. 
In this paper, we present SuperGlobal, a novel approach that exclusively employs global features for both stages, improving efficiency without sacrificing accuracy. 
SuperGlobal introduces key enhancements to the retrieval system, specifically focusing on the global feature extraction and reranking processes.
For extraction, we identify sub-optimal performance when the widely-used ArcFace loss and Generalized Mean (GeM) pooling methods are combined and propose several new modules to improve GeM pooling. %
In the reranking stage, we introduce a novel method to update the global features of the query and top-ranked images by only considering feature refinement with a small set of images, thus being very compute and memory efficient. %
Our experiments demonstrate substantial improvements compared to the state of the art in standard benchmarks.
Notably, on the Revisited Oxford+1M Hard dataset, our single-stage results improve by $7.1\%$, while our two-stage gain reaches $3.7\%$ with a strong $64,865\times$ speedup. 
Our two-stage system surpasses the current single-stage state-of-the-art by $16.3\%$, offering a scalable, accurate alternative for high-performing image retrieval systems with minimal time overhead.

\noindent Code: \url{https://github.com/ShihaoShao-GH/SuperGlobal}.

\end{abstract}

%% file: sections/intro_new.tex
\section{Introduction}
\label{sec:intro}

\begin{figure}[t]
\begin{center}
\includegraphics[width=1.0\linewidth]{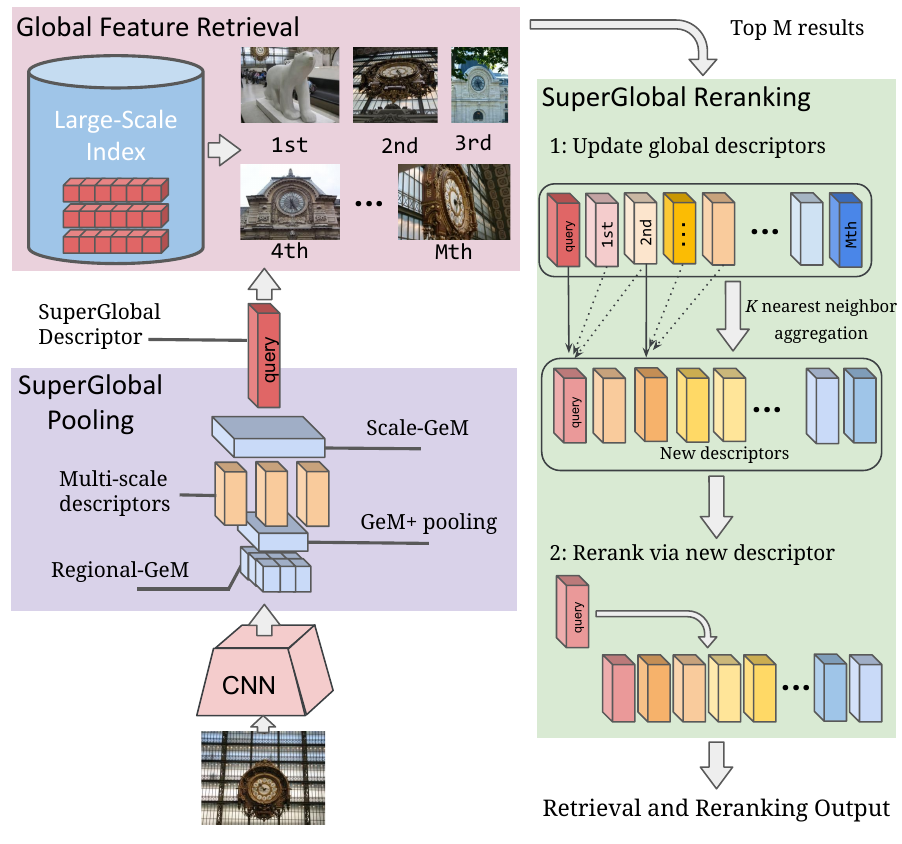}
\end{center}
\vspace{-10pt}
\caption{We introduce \textbf{SuperGlobal}, a novel method for image retrieval and reranking which relies solely on global image features.
SuperGlobal leverages several improvements to the Generalized Mean (GeM) pooling function, across regions and scales, as indicated in the purple box on the left.
Our reranking process, illustrated on the right green box, refines the global feature representation based on the query and top retrieved images to produce a more relevant set of results.
}
\label{fig:teaser}
\end{figure}
Image retrieval systems are tasked with searching large databases for visual contents similar to a query image.
Generally, the search process consists of two stages.
First, an efficient method sorts database images according to estimated high-level similarity to the query.
Then, in the reranking stage, the most relevant database images found in the first stage undergo a more comprehensive matching process against the query, to return an improved ranked list of results.

In modern implementations, the first stage is instantiated with deep learning-based global features, which has received substantial attention in the past few years \cite{radenovic2018fine,Revaud2019ICCV,Ng2020SOLARSL,yang2021dolg}.
The reranking stage is commonly executed via geometric matching of local image features \cite{Philbin07,Philbin2008,avrithis2014hough,cao2020unifying}, which provides information on the spatial consistency between the query and a given database image.

A recent trend in this area is on leveraging sophisticated matching processes at the reranking stage, \eg transformers \cite{Tan2021InstancelevelIR} or 4D correlation networks \cite{leecorrrelation2022}, which have led to remarkable improvements in the quality of retrieved results.
However, this has come at a significant cost, where reranking latency takes several seconds per query and requires more than $1$MB of memory per database image -- making these approaches challenging to scale to large repositories.
Our work directly tackles this limitation by proposing the first method fully based on global image features for both stages. 
In addition, we rethink pooling techniques and propose modules to improve global feature extraction. 
An overview of our method, SuperGlobal, is presented in Figure \ref{fig:teaser}.
More specifically, we introduce the following contributions.

\noindent\textbf{Contributions:}

\textbf{(1)} We propose improvements to the core global feature representation, based on enhanced pooling strategies.
We point out undesired training behavior when learning global features combining GeM pooling~\cite{radenovic2018fine} and ArcFace loss \cite{Deng2019ArcFaceAA}, and introduce a simple and effective solution to this problem with new pooling modules, including regional and multi-scale techniques.

\textbf{(2)} We introduce a very efficient and effective reranking process, based solely on global features, that is able to adapt the representation of the query and top-ranked database images on the fly in order to better estimate their similarities.
This method does away with any need for expensive local features and is inherently very scalable since the features used in both search stages are the same.

\textbf{(3)} Experiments on standard image retrieval benchmarks showcase the effectiveness of our methods, establishing new state-of-the-art results across the board.
We boost single-stage results on Revisited Oxford+1M Hard \cite{radenovic2018revisiting} by $7.1\%$.
But even more impressively, our simple reranking mechanism outperforms previous complex methods by $3.7\%$ on the same dataset, while being more than four orders of magnitude faster and requiring $170\times$ less memory.

%% file: sections/02_rw.tex
\section{Related Work}
\label{sec:rw}

\noindent\textbf{Image retrieval methods.}
Early work in image retrieval leveraged hand-crafted local features \cite{Lowe04distinctiveimage,bay2008speeded} as a core building block.
While some papers proposed to retrieve directly based on local features \cite{Mikolajczyk2002affine,Lowe04distinctiveimage,obdrzalek2005sublinear}, others used them to construct global representations, based on Bag-of-Words and similar techniques \cite{Sivic2003,jegou2010aggregating,jegou2012aggregating,tolias2015image}.
Modern systems have revisited these image retrieval techniques with deep learning based components, \eg, deep local feature-based retrieval \cite{Noh2017LargeScaleIR}, deep local feature aggregation \cite{Teichmann2019DetectToRetrieveER,Tolias2020LearningAA,weinzaepfel2022learning} or deep global feature modeling \cite{babenko2014neural,gordo2017end,Revaud2019ICCV,Ng2020SOLARSL,yang2021dolg}.
A recent survey in this area can be found in \cite{chen2022deep}.

\noindent\textbf{Global feature pooling.}
In particular, a critical aspect that has been studied for global feature learning is on how to properly pool contributions of image features from different regions into a single high-dimensional vector.
SPoC \cite{babenko2015iccv} proposed sum pooling of convolutional features, while \cite{razavian2016visual} introduced max pooling,
which was later approximated by integral max pooling in R-MAC \cite{tolias2015particular}.
Along a similar line, CroW \cite{kalantidis2015cross} introduced cross-dimensional weighted sum pooling.
NetVLAD \cite{arandjelovic2016netvlad} introduced an aggregation inspired by the VLAD method \cite{jegou2010aggregating}.
Generalized Mean (GeM) pooling \cite{radenovic2018fine} is today considered the state-of-the-art method in this area, being used in several recent papers \cite{leecorrrelation2022,cao2020unifying,yang2021dolg}.
A key contribution of our paper is to revisit global pooling methods, by pointing out the sub-optimal behavior of GeM when using a popular training loss, and by improving regional and multi-scale pooling.
Note that R-MAC \cite{tolias2015particular} had explored regional pooling, with max pooling over regions and sum pooling of these regional descriptors.
In contrast, we apply the more modern GeM pooling across regions and scales to achieve enhanced performance.

\noindent\textbf{Loss functions for image retrieval.}
Several types of loss functions have been developed to enhance instance-level discriminability, which is required in image retrieval systems.
Earlier work \cite{arandjelovic2016netvlad,gordo2017end,radenovic2018fine} in this area relied on ranking-based losses such as contrastive \cite{chopra2005learning} or triplet \cite{schroff2015facenet}.
More advanced ranking losses based on differentiable versions of Average Precision (AP) \cite{he2018local} have also demonstrated strong results \cite{Revaud2019ICCV}.
A recent trend is to leverage margin-based classification loss functions tuned to this problem, such as ArcFace \cite{Deng2019ArcFaceAA}, CosFace \cite{hao2018cosface} or CurricularFace \cite{huang2020curricularface} -- these have been adopted in image retrieval systems such as \cite{leecorrrelation2022,cao2020unifying,yang2021dolg}.
In this work, we point out a critical issue when these margin-based classification losses are coupled with GeM pooling -- which we fix with new pooling modules.

\noindent\textbf{Reranking for image retrieval.}
The reranking of image retrieval results has been traditionally accomplished by local feature matching and Geometric Verification (GV) \cite{Philbin07,Philbin2008,avrithis2014hough}, most often coupled to RANSAC \cite{Fischler1981}.
Modern deep local features \cite{Noh2017LargeScaleIR,cao2020unifying} have also been used in this manner.
A more recent trend is to employ heavier techniques for reranking, based on transformers \cite{Tan2021InstancelevelIR} or dense 4D correlation networks \cite{leecorrrelation2022}.
While achieving high performance, these incur substantial storage and compute costs due to the need to store local features and feed them through complex models.
Contrary to this trend, we propose a much simpler reranking technique where only global features are needed -- costing orders of magnitude less than the current state-of-the-art solution \cite{leecorrrelation2022} but still achieving higher accuracy.

%% file: sections/03_retrieval.tex
\section{Improving Global Features}
\label{sec:retrieval}

\subsection{Background} 
\label{subsec:background}

\noindent\textbf{GeM pooling.} Generalized Mean (GeM) pooling~\cite{radenovic2018fine} is a module that provides a generalized capability for feature aggregation. GeM pooling is widely adopted in ResNet \cite{He2016DeepRL} (RN for short) models for image retrieval~\cite{leecorrrelation2022,cao2020unifying,yang2021dolg}, followed by a fully-connected layer \cite{gordo2017end}, to whiten the aggregated representation. Formally, we denote the feature map from deep convolution layers
by $D\in\mathcal{R}^{H_d\times W_d\times C_d}$ and a fully-connected whitening layer with weights and bias as $W\in\mathcal{R}^{C_g\times C_d}$ 
and $b\in\mathcal{R}^{C_g}$, where $C_d$ and $C_g$ are the channel dimensions of the output from the convolution layer and global features, respectively.
These two components, GeM pooling and the whitening layer, produce the global feature $g\in\mathcal{R}^{C_g}$ by:

\begin{align}\label{eq:1}
    g = W \left(\frac{1}{H_d W_d} \sum_{h, w} D_{h,w}^p\right)^{1/p} + b,
\end{align}
where $p$ denotes the generalized mean power.

\noindent\textbf{SoftMax-based loss functions with margin penalties.} ArcFace~\cite{Deng2019ArcFaceAA} applies a geometric space penalty to expand the angular margin between different classes while gathering the same-class embedding to the center, therefore making it suitable for standard retrieval tasks \cite{yang2021dolg,cao2020unifying}. CurricularFace~\cite{huang2020curricularface} proposes to further improve angular margin losses by embedding curriculum training into the loss function and has the ability to adaptively adjust the relative importance of easy and hard samples during the course of training, which has been used in 
recent image retrieval work~\cite{leecorrrelation2022}.

\noindent
\textbf{Multi-scale inference.} Multi-scale inference is one of the commonly used methods to aggregate features from different scales to further improve the performance of image retrieval, previous papers ~\cite{cao2020unifying,yang2021dolg,leecorrrelation2022} commonly average the embeddings from different scales.

\subsection{Suboptimal GeM Pooling with Margin-based Losses}
\label{subsec:gem_arcface}

\begin{figure}[t]
\centerline{\includegraphics[width=0.97\linewidth]{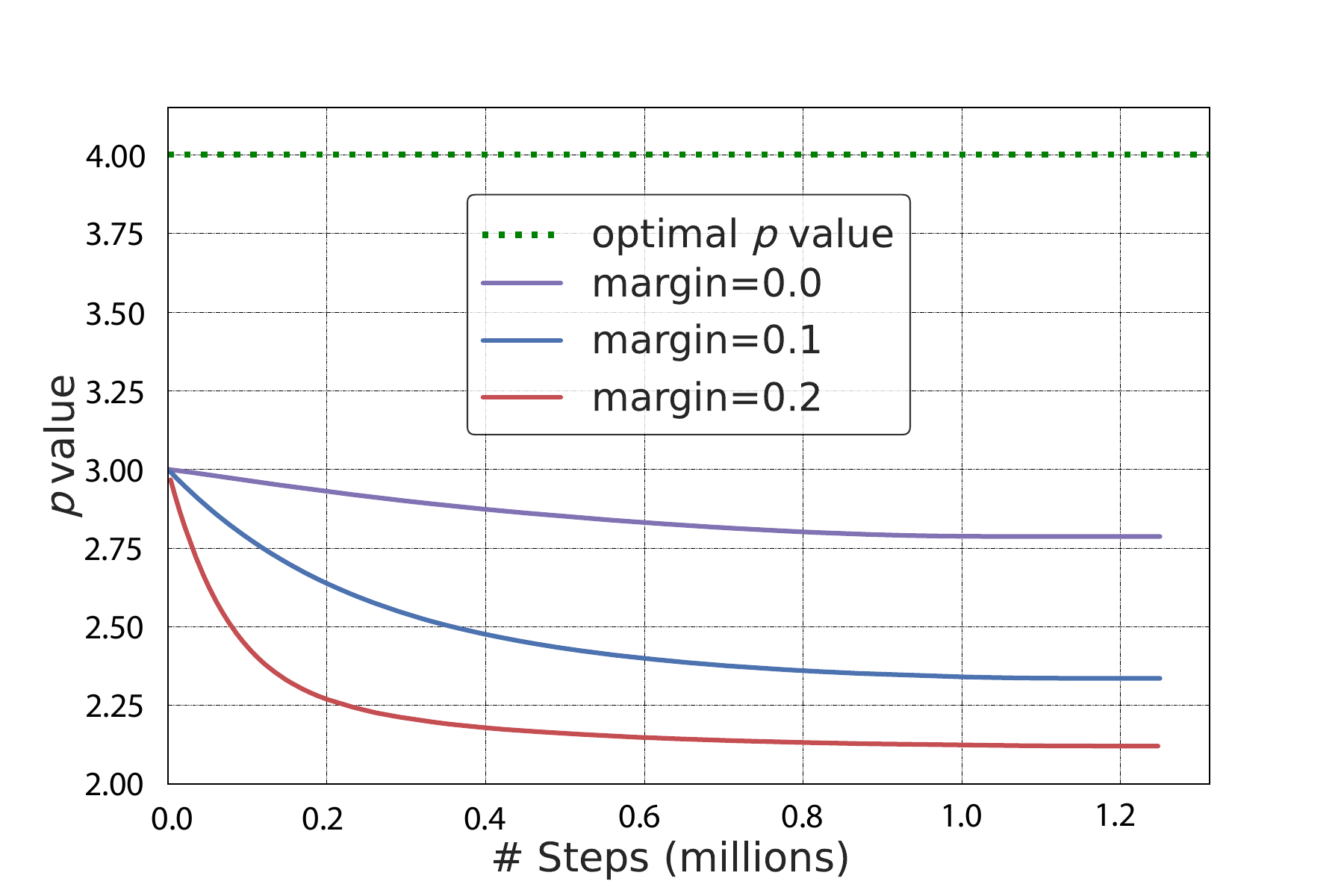}}
\vspace{-5pt}
\caption{\textbf{Trainable GeM pooling $p$ values during DELG training} for different ArcFace margin values, as compared to the optimal $p$ value (4.0). Note that larger margins cause $p$ to deviate further from its optimal value.}
\label{fig:p_margin}
\end{figure}

\begin{figure}[t]
\centerline{\includegraphics[width=1.0\linewidth]{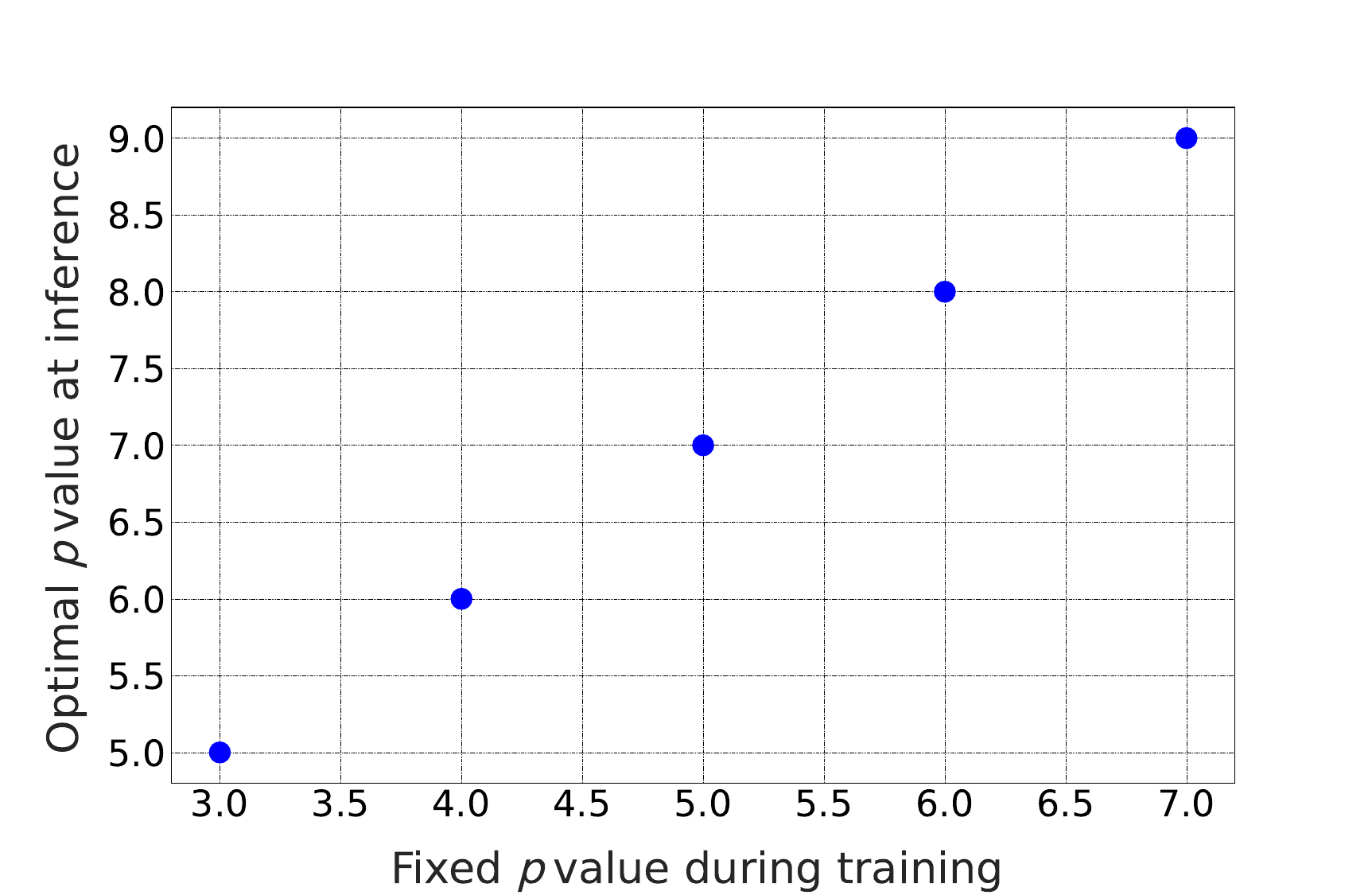}}
\vspace{-5pt}
\caption{\textbf{Optimal $p$ values at inference}, shown on the $y$-axis, for different fixed GeM pooling $p$ values during DELG training. We note that the optimal $p$ for inference is larger than the fixed $p$ used for training. 
}
\label{fig:fixed_p}
\end{figure}

We observe that combining CurricularFace or ArcFace loss with GeM pooling systematically causes the trainable $p$ value in GeM pooling to converge to a lower value w.r.t. its optimal value for image retrieval.
In Figure \ref{fig:p_margin}, we show such phenomena when training DELG \cite{cao2020unifying} with learnable $p$ values initialized to $3.0$, on GLDv2-clean \cite{weyand2020google}.
The optimal $p$ value in the test split of GLDv2-retrieval, found by simple grid search to inform the best possible value, is marked with a dotted green line (in this case, the optimal $p$ value for the converged model was found to be the same for these three runs).
During training, $p$ values keep decreasing from its original value, and are further away from the optimal value of $4.0$. 
Moreover,  higher angular margin causes $p$ to deviate further from its optimal value. 
In the cases of using fixed $p$ for GeM pooling during training, the optimal $p$ during inference could also be different from that in training. We examine the optimal $p$ values for inference by grid search in the test split of GLDv2-retrieval for different fixed $p$ values in training DELG models \cite{cao2020unifying} on GLDv2-clean \cite{weyand2020google} and find that optimal $p$ at inference is always higher, as shown in Figure \ref{fig:fixed_p}.
SOLAR~\cite{Ng2020SOLARSL} pointed out a similar phenomenon for models trained with the contrastive loss, while the underlying reason was not explained.%

In this section, we provide an intuition on the change of $p$ over the course of training based on our empirical study as follows. A high $p$ value generally forces a small portion of features to dominate the aggregation process. On the contrary, a lower $p$ leads to the opposite behavior: \eg, for $p = 1$, equivalent to average pooling, all features contribute equally. 
At the beginning of training, when the features are not refined, a lower $p$ value is preferred in order to gather more information; this is supported by the observation that $p$ goes down rapidly at the start of the training in Figure \ref{fig:p_margin}. In the later stage of training, when the features are further refined, focusing on critical features rather than all features may further improve the model performance. However, due to the decaying learning rate, $p$ value is not allowed to go up although higher $p$ may be preferred in this case. This aligns with Figure \ref{fig:p_margin}, where $p$ slowly converges later at training.

To conclude, we consistently observe that margin-based losses  lead to a sub-optimal $p$ value, resulting in the degradation of retrieval performance. This finding provides evidence of improvements and inspires future work for several state-of-the-art models, \eg CVNet \cite{leecorrrelation2022} which uses CurricularFace loss function and DELG \cite{cao2020unifying} with ArcFace loss. We introduce a set of modules specifically designed to optimize pooling for image retrieval in the following section.

\subsection{SuperGlobal Pooling} 
\label{subsec:revisited_pooling}
In this section, we revisit global feature pooling and propose three new modules to enhance retrieval capabilities: GeM+, Scale-GeM and Regional-GeM -- which are illustrated in the purple box in Figure~\ref{fig:teaser}.

\noindent
\textbf{GeM+.} As discussed in the previous subsection, GeM's $p$ value becomes sub-optimal with margin-based softmax losses.
Thus, we propose a method that starts by training with margin-based loss, then introduces a parameter tuning stage that will adjust $p$ in an efficient manner.
We find that in practice this tuning stage leads to the optimal value in a consistent manner for many datasets.
This approach is named GeM+ and seeks to find the optimal $p$ value of GeM pooling for image retrieval.

\noindent
\textbf{Regional-GeM.}
When adopting GeM for global pooling in image retrieval, we expect it to amplify discriminative information when aggregating the features to the final embedding. 
However, in addition to discriminative information at the global level, regional information such as object shape and arrangement can be important for distinguishing between different instances.
Such fine-grained details may not be captured robustly when simply pooling at the global level.
Therefore, besides using GeM pooling, we further adjust aggregation in order to incorporate regional information. We refer to this method as Regional-GeM.

We perform regional aggregation by adapting the $L_p$ pooling approach~\cite{gulcehre2014learned} to our network, with parameter $p_r$.
This can be viewed as a version of GeM pooling which acts on a regional level.
In this setup, activations from the feature map $D$ are aggregated in a convolutional manner, resulting in a new feature map, $M\in\mathcal{R}^{H_d\times W_d\times C_d}$.
We then combine $M$ and $D$ to produce a more robust feature map, obtaining an improved global feature as:

\begin{align}\label{eq:regional}
    g_r = W \left(\frac{1}{H_d W_d} \sum_{h, w} (\frac{M_{h,w} + D_{h,w}}{2})^p\right)^{1/p} + b.
\end{align}

With this formulation, we incorporate both regional information ($M_{h,w}$ produced by $L_p$ pooling with parameter $p_r$) and global information ($D_{h,w}$ produced by the original convolutional layer) in GeM pooling.
This module is integrated into our model without the need for additional training, leveraging the parameter $p$ obtained by the GeM+ process.

\noindent
\textbf{Scale-GeM.} 
Though averaging multi-scale features can be effective \cite{cao2020unifying,yang2021dolg,leecorrrelation2022}, a more generalized multi-scale aggregation may unlock larger retrieval gains.
With this motivation, we explore the application of GeM for enhanced multi-scale feature inference, and we refer to this as Scale-GeM.

GeM pooling can be applied before or after a fully-connected whitening layer. Our preliminary experiments applying it prior to projection yield sub-optimal performance, so we proceed by first extracting each scale's global feature according to Equation \ref{eq:regional}. 
Naively applying GeM pooling to such global features could fail due to the possible negative values in the features to be pooled. To address this issue, we consider a modified version of GeM designed for multi-scale inference as follows:
\begin{align}\label{eq:6}
    g_{ms} = \left( \frac{1}{N} \sum_{s=1}^N ( g_{s} + \zeta_s)^{p_{ms}} \right)^{1/{p_{ms}}} - \zeta_s,
\end{align}
\noindent where $\zeta_s=-min(g_s)$ denotes a shift of each scale's global feature $g_s$, $N$ denotes the number of scales and $p_{ms}$ is the multi-scale power parameter used in aggregation.

%% file: sections/04_reranking.tex
\section{Reranking with Global Features}
\label{sec:reranking}

\begin{figure*}[t]
\begin{center}
   \includegraphics[width=0.9\linewidth]{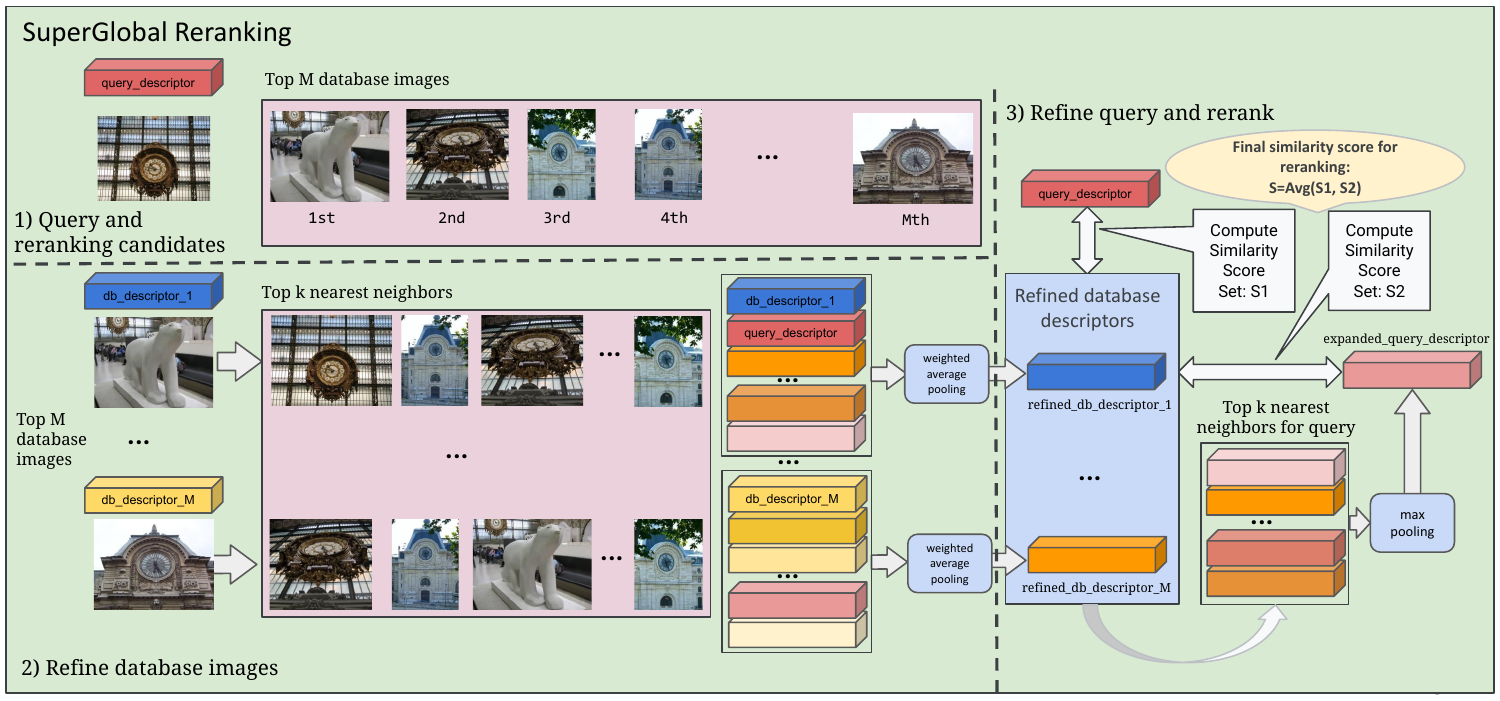}
\end{center}
   \caption{When reranking with global features, our system first performs feature aggregation for both query and top-$M$ retrieved images and then runs reranking via updated global features. More details can be found in Section \ref{subsec:reranking_details}.
   }
\label{fig:reranking}
\end{figure*}

\subsection{Refining Global Feature for Reranking}
\label{subsec:refine_global}

Robust image representations are critical for the accuracy of image retrieval. Combining the representations of similar images with that of the original image into an expanded representation that is then reissued as the query is a technique widely used to refine global features, generally leading to increased recall~\cite{gordo2020attention,arandjelovic2012}. 
Query expansion (QE)~\cite{gordo2020attention} is an example, as it replaces the original representation of the query image by its expanded version, which is then used to search better images in the database. 
On the other hand, database-side augmentation (DBA)~\cite{gordo2017end} is a method to apply QE to each image in the database. The key idea is that visually similar images are highly likely to contain the same object from different viewpoints and illumination conditions. 
Feature refinement with these images improves the robustness of the image representation.
It also emphasizes the key features of the object of interest, which further improves the representations. 
QE and DBA methods are very powerful but suffer from high cost: QE has to issue a new query against the entire database; DBA requires comparing all database images against each other, which can be infeasible in large scale.
Furthermore, adding a new image to the database with DBA requires querying it against the entire database. 

Reranking is usually conducted on the top-$M$ retrieved database images, where $M$ is much smaller than the database size -- making it feasible to apply feature refinement for each of these images on the fly, to then issue the updated query against the $M$ retrieved images with the updated representations.
Inspired by QE and DBA, our SuperGlobal reranking proposes a simple but effective method to aggregate information between the top-ranked images and the query, to update their image representations.
Unlike previous QE/DBA work~\cite{gordo2020attention,arandjelovic2012} that generally focuses on improving the features for a better recall, our work aims to refine the features for higher precision, since the reranking is performed on the top-$M$ results only. 

Selecting the candidate images for feature refinement may be challenging, since we don't have guarantees that they are actually relevant to the query. False positives may harm the expanded representation. Besides, the method to aggregate the features to reinforce the information shared between them and inject new information not available in the original representation is unclear. Our work addresses these challenges and proposes a reranking method only via refined global features, as described in the following.

\subsection{SuperGlobal Reranking}
\label{subsec:reranking_details}
SuperGlobal leverages GeM pooling to refine the global features for a given query and its top-$M$ retrieved images on the fly, and then reranks them via the updated descriptors, as illustrated in Figure~\ref{fig:reranking}. Different from previous studies and given that feature refinement runs at query time, the query image is also included as a candidate when refining the features for the database images. This helps to narrow the focus on query-specific feature refinement. The design details are discussed as follows. 

\noindent\textbf{Top-$K$ nearest neighbors as refinement candidates.}
For a given query image, we retrieve the top-$M$ images based on the global descriptors, where $M$ is a constant and typically below $1000$. Then for the query and the $M$ images, we fetch the top-$K$ nearest neighbors via global feature similarity, which are the candidates for the feature refinement, where $K$ is a constant and usually $K \leq 10$. 

\noindent\textbf{Feature refinement via GeM pooling.}
SuperGlobal reranking leverages GeM pooling for feature refinement. As previously mentioned, if there are false positives in the nearest neighbors, they may not be helpful but instead harmful to the expanded representation. Without strong Geometric Verification of local features to select true positives, the top-$K$ nearest neighbors could potentially contain false positives. SuperGlobal proposes effective strategies for database and query side separately as illustrated in Figure \ref{fig:reranking}.

For the database side, we propose a weighted pooling approach, with the global similarity score as the weight with additional multiplier factor $\beta$. After weighting the features, we demonstrate that applying average pooling ($p=1$) on top is the most effective for the database images. That is, $g_{dr}=(g_d + \sum^K_{i=1} (g_d\cdot g_i) \beta g_i)/(1+\sum^K_{i=1} (g_d\cdot g_i)\beta )$, where $g_d$ is the original global feature of the database image, $g_{dr}$ is the refined global feature we get and $g_i$ is the $i$-th most similar global feature to $g_d$. 

For the query side, we apply GeM pooling to the refined features of the top $K$ retrieved database images to produce an expanded global descriptor $g_{qe}$, and we find the optimal parameter $p$ is greater than 10, thus max pooling is applied (since, when $p \to \infty$, GeM pooling becomes max pooling). 
Both the original and the expanded descriptors of the query image are then used to compute the similarity scores for the final reranking, as follows.

\input{sections/tab_sota}

\noindent\textbf{Reranking with refined representations.}
Each query image possesses its original representation and the expanded representation. We compute the similarity scores $S1$ between each original descriptor $g_q$ and refined global descriptors $g_{dr}$ for each database image. We also compute another set of similarity scores $S2$ between the expanded query descriptor $g_{qe}$ and each $g_d$. In the end, we average $S1$ and $S2$ similarity scores for the final reranking. Given the fact that today's large-scale databases may contain billions of images, previous QE/DBA methods are much more costly compared to our approach, which has time complexity of $\mathcal{O}(M^2)$ and is extremely efficient at reranking.

%% file: sections/tab_sota.tex
\begin{table*}
\footnotesize
\addtolength{\tabcolsep}{-0.15em}
\centering
{%
\begin{tabular}{l c c c c c c c c c }
 \small \multirow{3}{*}{Method}  & \multicolumn{4}{c}{ \small Medium} && \multicolumn{4}{c}{ \small Hard} \\[+0.3em] \cmidrule[0.5pt]{2-5} \cmidrule[0.5pt]{7-10}
 & \multicolumn{1}{c}{\large \vphantom{M} \scriptsize $\mathcal{R}$Oxf } & \multicolumn{1}{c}{\scriptsize $\mathcal{R}$Oxf+1M} & \multicolumn{1}{c}{\scriptsize $\mathcal{R}$Par} & \multicolumn{1}{c}{\scriptsize $\mathcal{R}$Par+1M} && \multicolumn{1}{c}{\scriptsize $\mathcal{R}$Oxf} & \multicolumn{1}{c}{\scriptsize $\mathcal{R}$Oxf+1M} & \multicolumn{1}{c}{\scriptsize $\mathcal{R}$Par} & \multicolumn{1}{c}{\scriptsize $\mathcal{R}$Par+1M} \\
\midrule
\textbf{Global feature retrieval} \\
RN50-DELG \cite{cao2020unifying} & \num{73.6} & \num{60.6} & \num{85.7} & \num{68.6} &&  \num{51.0} & \num{32.7} & \num{71.5} & \num{44.4} \\
RN101-DELG \cite{cao2020unifying} & \num{76.3} & \num{63.7} & \num{86.6} & \num{70.6} &&  \num{55.6} & \num{37.5} & \num{72.4} & \num{46.9} \\
RN50-DOLG \cite{yang2021dolg} &   \num{80.5}  & \num[math-rm=\mathbf]{76.6} & \num{89.8} & \num{80.8} && \num{58.8} & \num{52.2} & \num{77.7} & \num{62.8} \\ 
RN101-DOLG \cite{yang2021dolg} & \num{81.5}  & \num{77.4} & \num{91.0}  & \num{83.3} && \num{61.1} & \num{54.8} & \num{80.3} & \num{66.7} \\
RN50-CVNet \cite{leecorrrelation2022} & \num{81.0} & \num{72.6} & \num{88.8} & \num{79.0} &&  \num{62.1} & \num{50.2} & \num{76.5} & \num{60.2} \\
RN101-CVNet \cite{leecorrrelation2022} & \num{80.2} & \num{74.0} & \num{90.3} & \num{80.6} && \num{63.1} & \num{53.7} & \num{79.1} & \num{62.2} \\
RN50-SuperGlobal (No reranking) \textbf{[ours]} & \num[math-rm=\mathbf]{83.9} & 74.7 & \num[math-rm=\mathbf]{90.5} & \num[math-rm=\mathbf]{81.3} &&  \num[math-rm=\mathbf]{67.7} & \num[math-rm=\mathbf]{53.6} & \num[math-rm=\mathbf]{80.3} & \num[math-rm=\mathbf]{65.2} \\
RN101-SuperGlobal (No reranking) \textbf{[ours]} & \num[math-rm=\mathbf,color=blue]{85.3} & \num[math-rm=\mathbf,color=blue]{78.8} & \num[math-rm=\mathbf,color=blue]{92.1} & \num[math-rm=\mathbf,color=blue]{83.9} &&  \num[math-rm=\mathbf,color=blue]{72.1} & \num[math-rm=\mathbf,color=blue]{61.9} & \num[math-rm=\mathbf,color=blue]{83.5} & \num[math-rm=\mathbf,color=blue]{69.1} \\
\midrule
\textbf{Global feature retrieval + Local feature reranking} \\
RN50-DELG (GV rerank top 100) \cite{cao2020unifying} & \num{78.3} & \num{67.2} & \num{85.7} & \num{69.6} &&  \num{57.9} & \num{43.6} & \num{71.0} & \num{45.7} \\
RN101-DELG (GV rerank top 100) \cite{cao2020unifying} & \num{81.2} & \num{69.1} & \num{87.2} & \num{71.5} &&  \num{64.0} & \num{47.5} & \num{72.8} & \num{48.7} \\
RN50-CVNet (Rerank top 400) \cite{leecorrrelation2022} & \num{87.9} & \num[math-rm=\mathbf]{80.7} & \num{90.5} & \num{82.4} &&  \num{75.6} & \num[math-rm=\mathbf]{65.1} & \num{80.2} & \num{67.3} \\
RN101-CVNet (Rerank top 400) \cite{leecorrrelation2022} & \num{87.2} & \num{81.9} & \num{91.2} & \num{83.8} &&  \num{75.9} & \num{67.4} & \num{81.1} & \num{69.3} \\
\midrule

\textbf{SuperGlobal retrieval and reranking} \\
RN50-SuperGlobal (Rerank top 400) \textbf{[ours]} & \num[math-rm=\mathbf]{88.8} & \num{80.0} & \num[math-rm=\mathbf]{92.0} & \num[math-rm=\mathbf]{83.4} &&  \num[math-rm=\mathbf]{77.1} & \num{64.2} & \num[math-rm=\mathbf]{84.4} & \num[math-rm=\mathbf]{68.7} \\
RN101-SuperGlobal (Rerank top 400) \textbf{[ours]}  & \num[math-rm=\mathbf,color=blue]{90.9} & \num[math-rm=\mathbf,color=blue]{84.4} & \num[math-rm=\mathbf,color=blue]{93.3} & \num[math-rm=\mathbf,color=blue]{84.9} &&  \num[math-rm=\mathbf,color=blue]{80.2} & \num[math-rm=\mathbf,color=blue]{71.1} & \num[math-rm=\mathbf,color=blue]{86.7} & \num[math-rm=\mathbf,color=blue]{71.4} \\
\end{tabular}
}
\caption{Results (\% mAP) on the $\mathcal{R}$Oxford and $\mathcal{R}$Paris datasets (and their large-scale versions $\mathcal{R}$Oxf+1M and $\mathcal{R}$Par+1M), with both Medium and Hard evaluation protocols. The best scores for RN50 and RN101, with and without reranking, are highlighted in \textbf{bold black} and \textbf{\textcolor{blue}{bold blue}}, respectively.}
\label{tab:sota}
\end{table*}

%% file: sections/05_experiments.tex
\section{Experiments}\label{sec:experiments}

\subsection{Experimental Setup}\label{subsec:experiment_detail}
\noindent\textbf{Common setting.} Our proposed methods can be applied to any model in a plug-in style. Here, we adopt our methods to the well-known structure CVNet \cite{leecorrrelation2022} with pre-trained weights downloaded from their GitHub repository. The modules we proposed in this paper are all implemented using TensorFlow \cite{tensorflow2015-whitepaper} and Pytorch \cite{paszkepytorch2019}. The training and inference are both conducted on four A100 GPUs with Intel$^\circledR$ Xeon
$^\circledR$ Gold 6330 CPU @ 2.00GHz.

\noindent\textbf{Estimating $p$, $p_r$ and $p_{ms}$.} We use $\mathcal{R}$Oxford 5k \cite{radenovic2018revisiting} as the tuning dataset to estimate the pooling parameters $p$, $p_r$ and $p_{ms}$ for GeM+, Regional-GeM and Scale-GeM, respectively, and show that the obtained values are sufficiently precise. Firstly, we run inference on the model and store the last feature map for every image. Then, we apply the different types of pooling, varying the pooling parameters, on the feature map for each image. To search for the optimal parameter, we begin by performing a grid search with a step size of 1 and monitor the mAP metric. 
We terminate the grid search if the mAP in the current iteration is smaller than the previous one. Then, we decrease the grid search step size to 0.1 and redo the previously mentioned steps. 
Once this procedure is completed, we obtain the values of $p=4.6$ and $p_r=2.5$.
For Scale-GeM, similar experimentation finds that $p_{ms} \to\infty$, \ie, max pooling over the multi-scale global features, leads to the best performance. 
These final obtained parameters are used for experimentation on all evaluation datasets.

\noindent\textbf{SuperGlobal reranking.} 
For reranking evaluations, we follow the same setting as CVNet and rerank the top 400 candidates in most experiments, \ie $M=400$.
Given that our method is drastically more efficient than CVNet, we also study the performance with larger $M$ in specific cases.
We pick $K=9$ for the reranking method and set $\beta=0.15$ for feature refinement described in Section \ref{subsec:reranking_details}.%

\noindent\textbf{ReLU adjustment.} During our reranking experimentation, following the same way as we explore the impact of $p$ in GeM pooling, we also revisit the ReLU activation \cite{krizhevskyimagenet2012} by considering a generalized version where the threshold is treated as a parameter denoted by $\alpha$, which is reduced to vanilla ReLU when $\alpha=0$. In the best setup, we set threshold $\alpha$ to $0.014$
for the first block and the joints between blocks.

\subsection{Evaluation Benchmarks}
We conduct our experiments on several well-established benchmarks. First, we use Oxford \cite{4270197} and Paris \cite{4587635} with revisited annotations \cite{radenovic2018revisiting}, referred to as $\mathcal{R}$Oxf and $\mathcal{R}$Par,
respectively. There are $4993$ ($6322$) database images in the $\mathcal{R}$Oxf ($\mathcal{R}$Par) dataset, and a
different query set for each, both with $70$ images. Large-scale results are further reported with the $\mathcal{R}$1M distractor
set \cite{radenovic2018revisiting}, which contains 1M images. In addition, we also report results on the Google Landmarks
dataset v2 (GLDv2) \cite{weyand2020google}, using the latest ground-truth version (2.1). GLDv2-retrieval has
$1129$ queries ($379$ validation and $750$ testing) and 762k database images.

\input{sections/tab_latency}
\input{sections/tab_gld_ext}

\subsection{Results}\label{subsec:results}
We compare different components of SuperGlobal against state-of-the-art models in Table~\ref{tab:sota}. We split the settings into three categories: (1) Global feature retrieval. (2) Global feature retrieval + Local feature reranking. (3) SuperGlobal retrieval and reranking. In addition to the comparisons of performance, to illustrate the efficiency of our method, we compare SuperGlobal against CVNet and DELG in the number of scales, reranking time and the peak memory consumption, and summarize the results in Table~\ref{tab:latency}.

Firstly, as seen from Table.~\ref{tab:sota}, SuperGlobal retrieval significantly outperforms existing models in single-stage retrieval. For instance, in setting (1), our methods (RN101-SuperGlobal without reranking) outperform the second best RN101-DOLG by a significant margin of +$7.1\%$ in Revisited Oxford+1M Hard. Under the retrieval then reranking paradigm in setting (2), SuperGlobal retrieval and reranking in setting (3) achieves +$3.7\%$ against the second best RN101-CVNet when reranking top 400 in Revisited Oxford+1M Hard. Moreover, SuperGlobal reranking is $64,865\times$ faster and requires $170\times$ less memory, as indicated by Table.~\ref{tab:latency}. Remarkably, our method, even including the reranking time, is almost as efficient as RN101-CVNet-Global with only almost zero overhead.
 
To evaluate our proposed method when reranking more candidates, we further conduct experiments on GLDv2-retrieval and show the results in Table.~\ref{tab:gld_ext}. First, by increasing the number of images in reranking, SuperGlobal achieves further performance improvements. Considering the significantly reduced latency and memory requirements of our method, SuperGlobal is capable of reranking many more images with the same compute budget. When increasing the reranking budget to top 800 or 1600 candidates, SuperGlobal shows superior performance compared with CVNet reranking (rerank top 100), while still being $16,216\times$ faster and $85\times$ more memory efficient.

\input{sections/tab_ablation}

\subsection{Ablation Study}\label{subsec:ablation_study}
To evaluate the contribution from each module, we conduct a detailed ablation on $\mathcal{R}$Oxf and $\mathcal{R}$Par, based on the RN101-CVNet pre-trained backbone. We sequentially add the modules one by one to examine whether they lead to a higher performance. 
Results are presented in Table~\ref{tab:ablation}. 
In summary, GeM+ contributes the most to the performance, while Regional-GeM and Scale-GeM make further improvements. Our finding of modifying ReLU also brings an additional +$1\%$ improvement.

\subsection{Qualitative Results}

\begin{figure*}[h]

\begin{center}
\includegraphics[width=1.0\linewidth]{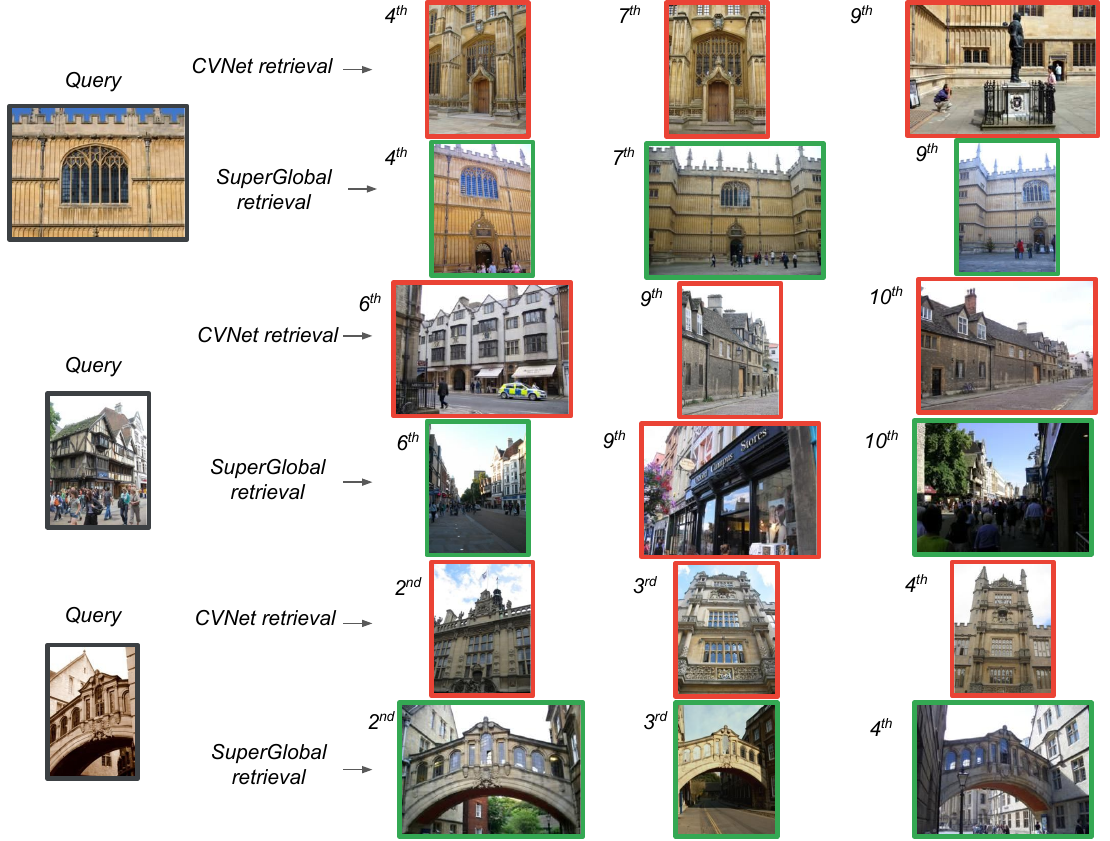}
\end{center}
\vspace{-10pt}
\caption{Examples of SuperGlobal retrieval and CVNet retrieval results on $\mathcal{R}$Oxf and $\mathcal{R}$Par dataset.}
\label{pic:global}
\end{figure*}

\noindent\textbf{Retrieval only.} In Figure \ref{pic:global}, we show images with different ranks retrieved from SuperGlobal and CVNet, in the absence of reranking. The ranking positions are selected such that SuperGlobal retrieves matching images (highlighted in green boxes) while CVNet doesn't (highlighted in red boxes). We observe that SuperGlobal pays more attention to the fine-grained details of the query image because of the updated pooling techniques proposed in this work.

\begin{figure*}[h]
\begin{center}
\includegraphics[width=1.0\linewidth]{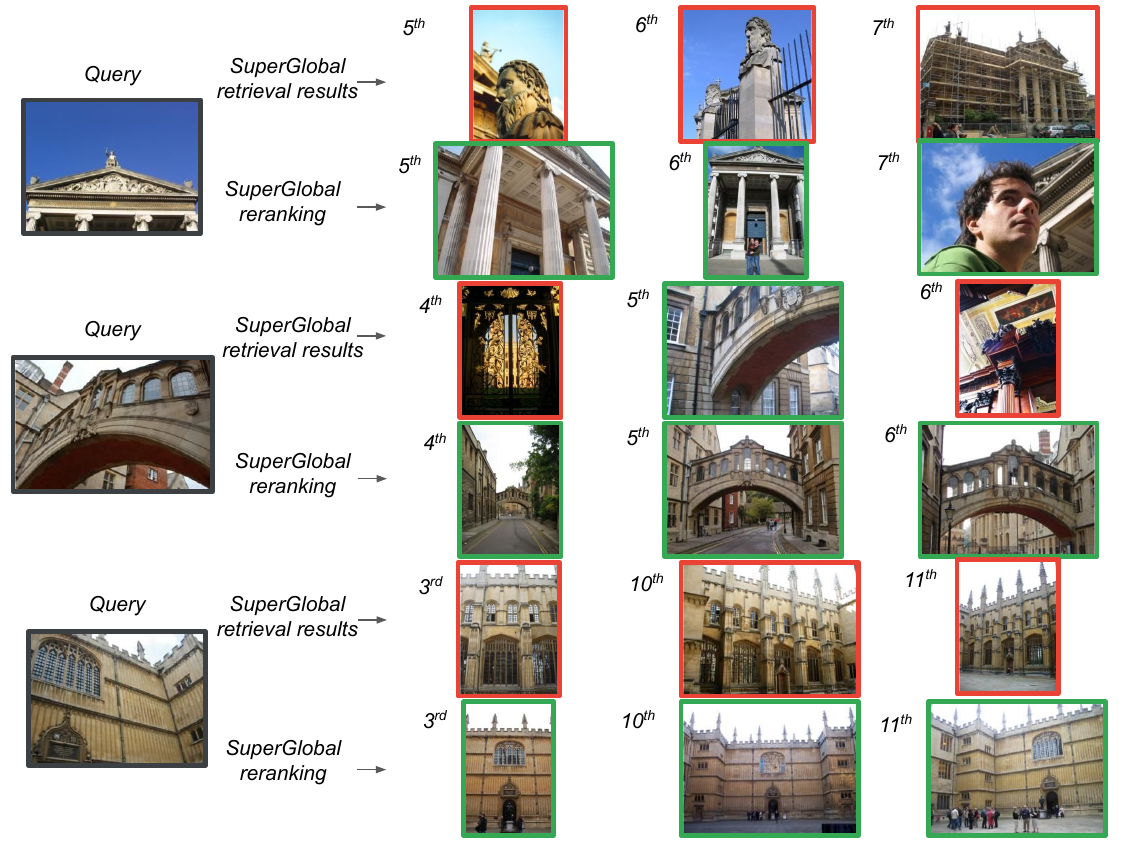}
\end{center}
\vspace{-10pt}
\caption{Examples of SuperGlobal retrieval and reranking results on $\mathcal{R}$Oxf and $\mathcal{R}$Par dataset.}
\label{pic:reranking_viz}
\end{figure*}

\noindent\textbf{Reranking.} Figure \ref{pic:reranking_viz} shows top results after SuperGlobal retrieval and reranking. The ranking positions are selected such that the reranked images (highlighted in green boxes) match the query whereas the retrieved images (highlighted in red boxes) do not. These examples show the additional improvement over single-stage SuperGlobal retrieval by applying SuperGlobal reranking, demonstrating the techniques in Section \ref{subsec:reranking_details} further refine the order of the top candidates.

\subsection{Local vs Global Feature Reranking}
\input{sections/tab_cvnetreranking}
SuperGlobal is proved to be significantly more efficient than CVNet reranking. For completeness, we perform experiments to examine whether conducting CVNet reranking on top of the SuperGlobal reranking results can further improve the performance. 
Table \ref{tab:cvnetrerank} shows that the results are not improved via CVNet reranking, except for a marginal improvement in $\mathcal{R}$Oxford Hard. This indicates that local and global feature reranking somehow overlap in the cases which they are able to improve, and our hypothesis for this is as follows. Global feature reranking combines features from visually similar images with diverse viewpoints and lighting conditions, leading to enhanced representation capability and robustness of the updated features. Therefore, global feature reranking could play a similar role as local feature reranking in retrieval systems and this might result in negligible gains when applying CVNet reranking on top of SuperGlobal. %

%% file: sections/tab_latency.tex
\begin{table*}
\footnotesize
\addtolength{\tabcolsep}{-0.15em}
\centering
{%
\begin{tabular}{l c c c c c c c c c c c}
 \small \multirow{2}{*}{Method} &  \multicolumn{2}{c}{ \small Multi-scale} && \multicolumn{1}{c}{ \small Extraction time} && \multicolumn{1}{c}{ \small Reranking time} && \multicolumn{2}{c}{ \small Memory (GB)}\\[+0.3em] \cmidrule[0.5pt]{2-10} \\
 &  \multicolumn{1}{c}{\small global } & \multicolumn{1}{c}{\small local } && (ms per image) && (ms on reranking top-400) && \multicolumn{1}{c}{\small $\mathcal{R}$Oxf} & \multicolumn{1}{c}{\small $\mathcal{R}$Par}  \\
\midrule
\textbf{Global features} \\
RN101-DELG \cite{cao2020unifying}  &  \num{3} & \num{7}&& 65 && \num[round-precision=2,round-mode=figures]{3.6e6} on 100  &&  \num{4.25} & \num{5.35}  \\
RN101-CVNet \cite{leecorrrelation2022}  &  \num{3} & \num{1}&& 65 && \num[round-precision=2,round-mode=figures]{2.4e4} on 400  &&  \num{27.02} & \num{33.55}  \\
RN101-CVNet$^Q$ \cite{leecorrrelation2022}  &  \num{3} & \num{1}&& 65 && \num[round-precision=2,round-mode=figures]{2.4e4} on 400 &&  \num{6.88} & \num{8.52}  \\
RN101-SuperGlobal (Ours)  &  \num{3} & - &&  65 && \num{0.37} on 400 &&  \num{0.04} & \num{0.05}  \\
\end{tabular}
}
\caption{Latency and Memory on the $\mathcal{R}$Oxford and $\mathcal{R}$Paris datasets . Extraction time measures the time needed for the model to produce global features. Reranking time measures the latency of the reranking stage after the global/local features are already computed. Memory usage measures the hardware memory required to store the features. %
}

\label{tab:latency}
\end{table*}

%% file: sections/tab_gld_ext.tex
\begin{table}[h]
\footnotesize
\addtolength{\tabcolsep}{-0.15em}
\centering
{%
\begin{tabular}{l c c c c}
Method & mAP@100 \\
\midrule
RN50-DELG retrieval& 24.1  \\
 + GV (Rerank top-100) & 24.3  \\
RN101-DELG retrieval& 26.0  \\
 + GV (Rerank top-100) & 26.8 \\
RN50-CVNet retrieval & 30.2  \\
 + CVNet reranking (Rerank top-100) & 32.4  \\
RN101-CVNet Retrieval & 32.5  \\
 + CVNet-reranking (Rerank top-100) & 34.9  \\
RN50-SuperGlobal retrieval \textbf{[ours]} & \num{31.1}  \\
 + SuperGlobal reranking (Rerank top-100) & 32.5 \\
 + SuperGlobal reranking (Rerank top-800) & \num[math-rm=\mathbf]{32.7}  \\
 + SuperGlobal reranking (Rerank top-1600) & 32.6  \\
RN101-SuperGlobal retrieval \textbf{[ours]} & \num{33.4}  \\
 + SuperGlobal reranking (Rerank top-100) & 34.6 \\
 + SuperGlobal reranking (Rerank top-800) & 34.9 \\
 + SuperGlobal reranking (Rerank top-1600) & \num[math-rm=\mathbf,color=blue]{35.0}  \\

\end{tabular}
}
\caption{\textbf{GLDv2-retrieval evaluation.} Experimental results (\% mAP@100) on GLDv2-retrieval \cite{weyand2020google}. The best scores are presented in \textbf{bold black} and \textcolor{blue}{\textbf{bold blue}} colors for each ResNet backbone.}
\label{tab:gld_ext}
\end{table}

%% file: sections/tab_ablation.tex
\begin{table*}[h!]
\footnotesize
\addtolength{\tabcolsep}{-0.15em}
\centering
{%
\begin{tabular}{l c c c c c c c c c }
 \small \multirow{3}{*}{Method} &   \small \multirow{3}{*}{GeM+} &  \small \multirow{3}{*}{Regional-GeM} &  \small \multirow{3}{*}{Scale-GeM} &  \small \multirow{3}{*}{ReLU} & \multicolumn{2}{c}{ \small Medium} && \multicolumn{2}{c}{ \small Hard} \\[+0.3em] \cmidrule[0.5pt]{6-7} \cmidrule[0.5pt]{9-10}
 &&&&& \multicolumn{1}{c}{\large \vphantom{M} \scriptsize $\mathcal{R}$Oxf } & \multicolumn{1}{c}{\scriptsize $\mathcal{R}$Par} && \multicolumn{1}{c}{\scriptsize $\mathcal{R}$Oxf} & \multicolumn{1}{c}{\scriptsize $\mathcal{R}$Par}  \\
\midrule
\textbf{Global features} \\
RN101-CVNet-Global \cite{leecorrrelation2022} &  \ding{56} & \ding{56}& \ding{56} & \ding{56} & \num{80.2} & \num{90.3}  &&  \num{63.1} & \num{79.1}  \\
RN101-CVNet-Global  &  \cmark & \ding{56}& \ding{56} & \ding{56} & \num{84.7} & \num{90.8}  &&  \num{69.6} & \num{81.1}  \\
RN101-CVNet-Global  &  \cmark & \cmark& \ding{56} & \ding{56} & \num{84.8} & \num{91.3}  &&  \num{70.6} & \num{81.9}  \\
RN101-CVNet-Global  &  \cmark & \cmark& \cmark & \ding{56} & \num{84.7} & \num{91.5}  &&  \num{71.1} & \num{82.5}  \\

RN101-CVNet-Global (SuperGlobal retrieval)  &  \cmark & \cmark& \cmark & \cmark & \num{85.3} & \num{92.1}  &&  \num{72.1} & \num{83.5}  \\

\end{tabular}
}
\caption{Results (\% mAP) on the $\mathcal{R}$Oxford 5k and $\mathcal{R}$Paris 6k datasets, with both Medium and Hard evaluation protocols. Note reranking is not applied in the evaluation.}
\label{tab:ablation}
\end{table*}

%% file: sections/tab_cvnetreranking.tex
\begin{table}[h!]
\footnotesize
\addtolength{\tabcolsep}{-0.15em}
\centering
{%
\begin{tabular}{l c c c c c c}
 \small \multirow{3}{*}{Method} &   \small \multirow{3}{*}{CVNet reranking}  & \multicolumn{2}{c}{ \small Medium} && \multicolumn{2}{c}{ \small Hard} \\[+0.3em] \cmidrule[0.5pt]{3-4} \cmidrule[0.5pt]{6-7}
 && \multicolumn{1}{c}{\large \vphantom{M} \scriptsize $\mathcal{R}$Oxf } & \multicolumn{1}{c}{\scriptsize $\mathcal{R}$Par} && \multicolumn{1}{c}{\scriptsize $\mathcal{R}$Oxf} & \multicolumn{1}{c}{\scriptsize $\mathcal{R}$Par}  \\
\midrule

\multirow{2}{*}{SuperGlobal} &   \ding{56} & \num{90.9} & \num{93.3}  &&  \num{80.2} & \num{86.7}  \\
  &   \checkmark & \num{90.9} & \num{91.9}  &&  \num{81.0} & \num{79.6}  \\

\end{tabular}

}

\caption{Results (\% mAP) of conducting CVNet reranking on top of our SuperGlobal reranking results on the $\mathcal{R}$Oxford and $\mathcal{R}$Paris datasets, with both Medium and Hard evaluation protocols.} %
\label{tab:cvnetrerank}
\end{table}

%% file: sections/06_conclusion.tex
\section{Conclusions}
\label{sec:conclusions}

In this paper, we propose a novel image retrieval system, SuperGlobal, which consists of various modules to refine global features for image retrieval and reranking.
All of our proposed methods can be plugged into other existing models, and are easy to implement.
For global feature refinement, we proposed improved pooling techniques by better training, besides leveraging regional and multi-scale components.
In contrast to conventional expensive reranking systems, we devise a strategy that requires only global features, delivering much improved performance while being four orders of magnitude more efficient.
This paper marks a first solution to the retrieval and reranking problems relying on a single global image feature.
We hope this will spur further research around this direction, to enable continued improvements to the scalabity of these systems.

%% file: sections/S4.tex
\section{Reranking more candidates in SuperGlobal}
\label{sec:rerank_supp}

\input{sections/tab_sota_supp}

This section serves as an extension to Section \textcolor{red}{5.3} and further evaluates using SuperGlobal with additional candidates on Revisited Oxford 5k (+1M) and Revisited Paris 6k (+1M) \cite{radenovic2018revisiting}. 
As presented in Table \ref{tab:sota_supp}, SuperGlobal achieves further performance improvements when reranking additional images. It significantly outperforms CVNet reranking by $6.9\%$ on Revisited Oxford+1M Hard, and surpasses SuperGlobal (rerank top 400) by $3.2\%$ in the same dataset. Even with additional candidates, SuperGlobal (rerank top 1600) enjoys significant latency gains over CVNet reranking (rerank top 400).

%% file: sections/tab_sota_supp.tex
\begin{table*}[t]
\footnotesize
\addtolength{\tabcolsep}{-0.15em}
\centering
{%
\begin{tabular}{l c c c c c c c c c }
 \small \multirow{3}{*}{Method}  & \multicolumn{4}{c}{ \small Medium} && \multicolumn{4}{c}{ \small Hard} \\[+0.3em] \cmidrule[0.5pt]{2-5} \cmidrule[0.5pt]{7-10}
 & \multicolumn{1}{c}{\large \vphantom{M} \scriptsize $\mathcal{R}$Oxf } & \multicolumn{1}{c}{\scriptsize $\mathcal{R}$Oxf+1M} & \multicolumn{1}{c}{\scriptsize $\mathcal{R}$Par} & \multicolumn{1}{c}{\scriptsize $\mathcal{R}$Par+1M} && \multicolumn{1}{c}{\scriptsize $\mathcal{R}$Oxf} & \multicolumn{1}{c}{\scriptsize $\mathcal{R}$Oxf+1M} & \multicolumn{1}{c}{\scriptsize $\mathcal{R}$Par} & \multicolumn{1}{c}{\scriptsize $\mathcal{R}$Par+1M} \\
\midrule
\textbf{Global feature retrieval} \\
RN50-DELG \cite{cao2020unifying} & \num{73.6} & \num{60.6} & \num{85.7} & \num{68.6} &&  \num{51.0} & \num{32.7} & \num{71.5} & \num{44.4} \\
RN101-DELG \cite{cao2020unifying} & \num{76.3} & \num{63.7} & \num{86.6} & \num{70.6} &&  \num{55.6} & \num{37.5} & \num{72.4} & \num{46.9} \\
RN50-DOLG \cite{yang2021dolg} &   \num{80.5}  & \num[math-rm=\mathbf]{76.6} & \num{89.8} & \num{80.8} && \num{58.8} & \num{52.2} & \num{77.7} & \num{62.8} \\ 
RN101-DOLG \cite{yang2021dolg} & \num{81.5}  & \num{77.4} & \num{91.0}  & \num{83.3} && \num{61.1} & \num{54.8} & \num{80.3} & \num{66.7} \\
RN50-CVNet \cite{leecorrrelation2022} & \num{81.0} & \num{72.6} & \num{88.8} & \num{79.0} &&  \num{62.1} & \num{50.2} & \num{76.5} & \num{60.2} \\
RN101-CVNet \cite{leecorrrelation2022} & \num{80.2} & \num{74.0} & \num{90.3} & \num{80.6} && \num{63.1} & \num{53.7} & \num{79.1} & \num{62.2} \\
RN50-SuperGlobal \textbf{[ours]} & \num[math-rm=\mathbf]{83.9} & \num{74.7} & \num[math-rm=\mathbf]{90.5} & \num[math-rm=\mathbf]{81.3} &&  \num[math-rm=\mathbf]{67.7} & \num[math-rm=\mathbf]{53.6} & \num[math-rm=\mathbf]{80.3} & \num[math-rm=\mathbf]{65.2} \\
RN101-SuperGlobal \textbf{[ours]} & \num[math-rm=\mathbf,color=blue]{85.3} & \num[math-rm=\mathbf,color=blue]{78.8} & \num[math-rm=\mathbf,color=blue]{92.1} & \num[math-rm=\mathbf,color=blue]{83.9} &&  \num[math-rm=\mathbf,color=blue]{72.1} & \num[math-rm=\mathbf,color=blue]{61.9} & \num[math-rm=\mathbf,color=blue]{83.5} & \num[math-rm=\mathbf,color=blue]{69.1} \\
\midrule
\textbf{Global feature retrieval + Local feature reranking} \\
RN50-DELG (GV rerank top 100) \cite{cao2020unifying} & \num{78.3} & \num{67.2} & \num{85.7} & \num{69.6} &&  \num{57.9} & \num{43.6} & \num{71.0} & \num{45.7} \\
RN101-DELG (GV rerank top 100) \cite{cao2020unifying} & \num{81.2} & \num{69.1} & \num{87.2} & \num{71.5} &&  \num{64.0} & \num{47.5} & \num{72.8} & \num{48.7} \\
RN50-CVNet (Rerank top 400) \cite{leecorrrelation2022} & \num{87.9} & \num{80.7} & \num{90.5} & \num{82.4} &&  \num{75.6} & \num{65.1} & \num{80.2} & \num{67.3} \\
RN101-CVNet (Rerank top 400) \cite{leecorrrelation2022} & \num{87.2} & \num{81.9} & \num{91.2} & \num{83.8} &&  \num{75.9} & \num{67.4} & \num{81.1} & \num{69.3} \\
\midrule

\textbf{SuperGlobal feature retrieval and reranking} \\
RN50-SuperGlobal (Rerank top 400) \textbf{[ours]} & \num{88.8} & \num{80.0} & \num{92.0} & \num{83.4} &&  \num{77.1} & \num{64.2} & \num{84.4} & \num{68.7} \\
RN101-SuperGlobal (Rerank top 400) \textbf{[ours]}  & \num{90.9} & \num{84.4} & \num{93.3} & \num{84.9} &&  \num{80.2} & \num{71.1} & \num{86.7} & \num{71.4} \\
RN50-SuperGlobal (Rerank top 800) \textbf{[ours]} & \num[math-rm=\mathbf]{88.9} & \num{81.3} & \num{93.0} & \num{85.4} &&  \num[math-rm=\mathbf]{77.4} & \num{67.0} & \num{86.2} & \num[math-rm=\mathbf]{75.4} \\
RN101-SuperGlobal (Rerank top 800) \textbf{[ours]}  & \num[math-rm=\mathbf,color=blue]{91.2} & \num{85.5} & \num{94.1} & \num{86.5} &&  \num[math-rm=\mathbf,color=blue]{80.7} & \num{73.5} & \num{88.2} & \num{74.6} \\
RN50-SuperGlobal (Rerank top 1600) \textbf{[ours]} & \num{88.9} & \num[math-rm=\mathbf]{82.0} & \num[math-rm=\mathbf]{93.3} & \num[math-rm=\mathbf]{86.8} &&  \num{76.9} & \num[math-rm=\mathbf]{68.2} & \num[math-rm=\mathbf]{86.4} & \num{75.0} \\
RN101-SuperGlobal (Rerank top 1600) \textbf{[ours]}  & \num{91.2} & \num[math-rm=\mathbf,color=blue]{85.9} & \num[math-rm=\mathbf,color=blue]{94.2} & \num[math-rm=\mathbf,color=blue]{87.7} &&  \num{80.6} & \num[math-rm=\mathbf,color=blue]{74.3} & \num[math-rm=\mathbf,color=blue]{88.4} & \num[math-rm=\mathbf,color=blue]{77.0} \\
\end{tabular}
}
\caption{\textbf{Comparison to the state-of-the-art methods in image retrieval tasks.} Results (\% mAP) on the $\mathcal{R}$Oxford and $\mathcal{R}$Paris datasets\cite{radenovic2018revisiting} (and their large-scale versions $\mathcal{R}$Oxf+1M and $\mathcal{R}$Par+1M), with both Medium and Hard evaluation protocols. Our SuperGlobal retrieval framework outperforms state-of-the-art image retrieval
methods by a large margin for every measure.  The best scores for RN50 and RN101, with and without reranking, are highlighted in \textbf{bold black} and \textbf{\textcolor{blue}{bold blue}}, respectively.}%
\label{tab:sota_supp}
\end{table*}

%% file: sections/S5.tex
\section{Parameter study for each module}
\label{sec:retrieval_rerank}

Our method consists of several modules that are sensitive to the choice of parameter. Therefore, one important aspect of our work is to seek the optimal values for the core parameters of each component. 
In this section, we verify the validity of these values by conducting grid searches on different parameter values. When appropriate, we present two digits after the decimal due to the minor differences in values.

\input{sections/tab_gem+_supp}

\input{sections/tab_scalegem_supp}

\input{sections/tab_regionalgem_supp}

\input{sections/tab_relu_supp}

\noindent\textbf{$p$ for GeM+.}
As mentioned in Section \textcolor{red}{5.1}, we used $\mathcal{R}$Oxford 5k \cite{radenovic2018revisiting} to estimate $p$ and obtain the value of $4.6$ for GeM+. We show the results of different $p$ values in Table \ref{tab:gem+_supp} to verify that our $p$ is optimal.

\noindent\textbf{$p_{ms}$ for Scale-GeM.}
Here we perform grid search to explore the influence of $p_{ms}$ in Scale-GeM. The results are detailed in Table \ref{tab:scalegem_supp}.

\noindent\textbf{$p_r$ for Regional-GeM.}
Regional GeM consists of $L_p$ pool and GeM+. Table \ref{tab:regionalgem_supp} shows how the $p_r$ value of $L_p$ pool affects retrieval performance.
 
\noindent\textbf{ReLU threshold.}
Here we present the study of the relationship between the threshold $\alpha$ of ReLU and retrieval performance. 
We conduct grid search to investigate the optimal $\alpha$ and summarize the results in Table \ref{tab:reluthres_supp}.

%% file: sections/tab_gem+_supp.tex
\begin{table}[h!]
\footnotesize
\addtolength{\tabcolsep}{-0.15em}
\centering
{%\scriptsize
\begin{tabular}{l c c c c c c}
%\toprule
 \small \multirow{3}{*}{Method} &   \small \multirow{3}{*}{$p$}  & \multicolumn{2}{c}{ \small Medium} && \multicolumn{2}{c}{ \small Hard} \\[+0.3em] \cmidrule[0.5pt]{3-4} \cmidrule[0.5pt]{6-7}
 && \multicolumn{1}{c}{\large \vphantom{M} \scriptsize $\mathcal{R}$Oxf } & \multicolumn{1}{c}{\scriptsize $\mathcal{R}$Par} && \multicolumn{1}{c}{\scriptsize $\mathcal{R}$Oxf} & \multicolumn{1}{c}{\scriptsize $\mathcal{R}$Par}  \\
\midrule
%\midrule
% ResNet50-DELG \cite{cao2020unifying} &   \num{43.3} & \num{24.2}  & \num{58.0}  & \num{29.9} &&  \num{17.1} & \num{9.4} & \num{29.7} & \num{8.4} \\
% ResNet101=DELG \cite{cao2020unifying} &     \num{61.9} &  \num{42.6} & \num{69.3}  & \num{45.4} && \num{33.7} & \num{19.0} & \num{44.3} & \num{19.1} \\ 
\multirow{5}{*}{SuperGlobal}  &   4.2 & \num{90.9} & \num{93.4}  &&  \num{80.1} & \num{86.7}  \\
&  4.4 & \num{90.9} & \num{93.4}  &&  \num{80.1} & \num{86.8}  \\

&  4.6 & \num{90.9} & \num{93.3}  &&  \num{80.2} & \num{86.7}  \\

&  4.8 & \num{91.0} & \num{92.3}  &&  \num{80.3} & \num{86.7}  \\

&  5.0 & \num{90.8} & \num{93.3}  &&  \num{80.0} & \num{86.7}  \\

\end{tabular}

}

% \vspace{-5pt}
\caption{Results (\% mAP) of conducting grid search on different GeM+ $p$ values on the $\mathcal{R}$Oxford and $\mathcal{R}$Paris datasets \cite{radenovic2018revisiting}, with both Medium and Hard evaluation protocols.}
\label{tab:gem+_supp}
\end{table}

%% file: sections/tab_scalegem_supp.tex
\begin{table}[h]
\footnotesize
\addtolength{\tabcolsep}{-0.15em}
\centering
{%
\begin{tabular}{l c c c c c c}
 \small \multirow{3}{*}{Method} &   \small \multirow{3}{*}{$p_{ms}$}  & \multicolumn{2}{c}{ \small Medium} && \multicolumn{2}{c}{ \small Hard} \\[+0.3em] \cmidrule[0.5pt]{3-4} \cmidrule[0.5pt]{6-7}
 && \multicolumn{1}{c}{\large \vphantom{M} \scriptsize $\mathcal{R}$Oxf } & \multicolumn{1}{c}{\scriptsize $\mathcal{R}$Par} && \multicolumn{1}{c}{\scriptsize $\mathcal{R}$Oxf} & \multicolumn{1}{c}{\scriptsize $\mathcal{R}$Par}  \\
\midrule
\multirow{6}{*}{SuperGlobal}  &   1.0 & \num{89.5} & \num{93.1}  &&  \num{77.8} & \num{86.2}  \\
&  1.5 & \num{89.5} & \num{93.1}  &&  \num{77.9} & \num{86.2}  \\

&  2.0 & \num{89.7} & \num{93.1}  &&  \num{78.1} & \num{86.2}  \\

&  2.5 & \num{90.4} & \num{93.1}  &&  \num{79.1} & \num{86.2}  \\

&  3.0 & \num{90.6} & \num{93.1}  &&  \num{79.3} & \num{86.3}  \\

&  $+\infty$ & \num{90.9} & \num{93.3}  &&  \num{80.2} & \num{86.7}  \\

\end{tabular}

}

\caption{Results (\% mAP) of conduct grid search on different Scale-GeM $p_{ms}$ values on the $\mathcal{R}$Oxford and $\mathcal{R}$Paris datasets \cite{radenovic2018revisiting}, with both Medium and Hard evaluation protocols.}
\label{tab:scalegem_supp}
\end{table}

%% file: sections/tab_regionalgem_supp.tex
\begin{table}[h!]
\footnotesize
\addtolength{\tabcolsep}{-0.15em}
\centering
{%
\begin{tabular}{l c c c c c c}
 \small \multirow{3}{*}{Method} &   \small \multirow{3}{*}{$p_r$}  & \multicolumn{2}{c}{ \small Medium} && \multicolumn{2}{c}{ \small Hard} \\[+0.3em] \cmidrule[0.5pt]{3-4} \cmidrule[0.5pt]{6-7}
 && \multicolumn{1}{c}{\large \vphantom{M} \scriptsize $\mathcal{R}$Oxf } & \multicolumn{1}{c}{\scriptsize $\mathcal{R}$Par} && \multicolumn{1}{c}{\scriptsize $\mathcal{R}$Oxf} & \multicolumn{1}{c}{\scriptsize $\mathcal{R}$Par}  \\
\midrule
\multirow{5}{*}{SuperGlobal}  &   2.0 & \num{90.9} & \num{93.3}  &&  \num{80.2} & \num{86.7}  \\
&  2.2 & \num{90.9} & \num{93.3}  &&  \num{80.2} & \num{86.7}  \\

&  2.4  & \num{90.9} & \num{93.3}  &&  \num{80.2} & \num{86.7}  \\

&  2.6 & \num{90.8} & \num{93.3}  &&  \num{80.0} & \num{86.7}  \\

&  2.8 & \num{90.8} & \num{93.4}  &&  \num{80.0} & \num{86.7}  \\

\end{tabular}

}

\caption{Results (\% mAP) of conducting grid search on different Regional-GeM $p_r$ values on the $\mathcal{R}$Oxford and $\mathcal{R}$Paris datasets \cite{radenovic2018revisiting}, with both Medium and Hard evaluation protocols.}
\label{tab:regionalgem_supp}
\end{table}

%% file: sections/tab_relu_supp.tex
\begin{table}[h!]
\footnotesize
\addtolength{\tabcolsep}{-0.15em}
\centering
{%
\begin{tabular}{l c c c c c c}
 \small \multirow{3}{*}{Method} &   \small \multirow{3}{*}{$\alpha$}  & \multicolumn{2}{c}{ \small Medium} && \multicolumn{2}{c}{ \small Hard} \\[+0.3em] \cmidrule[0.5pt]{3-4} \cmidrule[0.5pt]{6-7}
 && \multicolumn{1}{c}{\large \vphantom{M} \scriptsize $\mathcal{R}$Oxf } & \multicolumn{1}{c}{\scriptsize $\mathcal{R}$Par} && \multicolumn{1}{c}{\scriptsize $\mathcal{R}$Oxf} & \multicolumn{1}{c}{\scriptsize $\mathcal{R}$Par}  \\
\midrule
\multirow{5}{*}{SuperGlobal}  &   0.012 & \num{90.7} & \num{93.3}  &&  \num{79.8} & \num{86.7}  \\
&  0.014 & \num{90.9} & \num{93.3}  &&  \num{80.2} & \num{86.7}  \\

&  0.016 & \num{90.9} & \num{93.4}  &&  \num{80.0} & \num{86.7}  \\

&  0.018 & \num{90.9} & \num{93.4}  &&  \num{80.2} & \num{86.7}  \\

&  0.020 & \num{90.8} & \num{93.3}  &&  \num{80.1} & \num{86.6}  \\

\end{tabular}

}

\caption{Results (\% mAP) of conducting grid search on different ReLU threshold on the $\mathcal{R}$Oxford and $\mathcal{R}$Paris datasets \cite{radenovic2018revisiting}, with both Medium and Hard evaluation protocols.}
\label{tab:reluthres_supp}
\end{table}

%% file: sections/S6.tex
\input{sections/tab_delg_new_supp}
\input{sections/tab_cvnet_supp}

\section{Combining SuperGlobal with other state-of-the-art models.}
SuperGlobal can easily be adopted to existing retrieval methods for further improvements. Table \ref{tab:delg_new_supp} demonstrates that adopting SuperGlobal modules (GeM+, Scale-GeM, and Regional-GeM) and further performing SuperGlobal reranking on the DELG \cite{cao2020unifying} pretrained weights outperforms CVNet reranking \cite{leecorrrelation2022}.

%% file: sections/tab_delg_new_supp.tex
\begin{table*}[h!]
\footnotesize
\addtolength{\tabcolsep}{-0.15em}
\centering
{%
\begin{tabular}{l c c c c c}
 \small \multirow{3}{*}{Method} &    \multicolumn{2}{c}{ \small Medium} && \multicolumn{2}{c}{ \small Hard} \\[+0.3em] \cmidrule[0.5pt]{2-3} \cmidrule[0.5pt]{5-6}
 & \multicolumn{1}{c}{\large \vphantom{M} \scriptsize $\mathcal{R}$Oxf } & \multicolumn{1}{c}{\scriptsize $\mathcal{R}$Par} && \multicolumn{1}{c}{\scriptsize $\mathcal{R}$Oxf} & \multicolumn{1}{c}{\scriptsize $\mathcal{R}$Par}  \\
\midrule

{RN101-DELG\cite{cao2020unifying}}  & \num{76.3} & \num{86.6}  &&  \num{55.6} & \num{72.4}  \\
{RN101-DELG+SuperGlobal pooling [one-stage]}  & \num{80.0} & \num{90.6}  &&  \num{60.0} & \num{79.8}  \\
{RN101-DELG+SuperGlobal pooling and reranking (top 400)} & \num{88.4} & \num{93.1}  &&  \num{77.3} & \num{86.8}  \\

\end{tabular}

}

\caption{Results (\% mAP) of adopting SuperGlobal to make further improvement on DELG \cite{cao2020unifying} on the $\mathcal{R}$Oxford and $\mathcal{R}$Paris datasets \cite{radenovic2018revisiting}, with both Medium and Hard evaluation protocols. }
\label{tab:delg_new_supp}
\end{table*}

%% file: sections/tab_cvnet_supp.tex
\begin{table*}[h!]
\footnotesize
\addtolength{\tabcolsep}{-0.15em}
\centering
{%
\begin{tabular}{l c c c c c c}
 \small \multirow{3}{*}{Method} &   \small \multirow{3}{*}{SuperGlobal (Rerank top 400)}  & \multicolumn{2}{c}{ \small Medium} && \multicolumn{2}{c}{ \small Hard} \\[+0.3em] \cmidrule[0.5pt]{3-4} \cmidrule[0.5pt]{6-7}
 && \multicolumn{1}{c}{\large \vphantom{M} \scriptsize $\mathcal{R}$Oxf } & \multicolumn{1}{c}{\scriptsize $\mathcal{R}$Par} && \multicolumn{1}{c}{\scriptsize $\mathcal{R}$Oxf} & \multicolumn{1}{c}{\scriptsize $\mathcal{R}$Par}  \\
\midrule

\multirow{2}{*}{RN101-CVNet-Global\cite{leecorrrelation2022}} &   \ding{56} & \num{80.2} & \num{90.3}  &&  \num{63.1} & \num{79.1}  \\
  &   \checkmark & \num{83.7} & \num{91.6}  &&  \num{68.6} & \num{82.5}  \\

\end{tabular}

}

\caption{Results (\% mAP) of adopting SuperGlobal (only reranking) on CVNet-Global \cite{leecorrrelation2022} on the $\mathcal{R}$Oxford and $\mathcal{R}$Paris datasets \cite{radenovic2018revisiting}, with both Medium and Hard evaluation protocols.} %
\label{tab:cvnet_supp}
\end{table*}

%% file: sections/S7.tex
\section{Generalizing SuperGlobal reranking}
SuperGlobal proposes the idea to rerank by further improving global feature of images via feature aggregation. This idea can be generalized when combined with other global features, \eg, DELG-Global \cite{cao2020unifying}, DOLG \cite{yang2021dolg} or CVNet-Global \cite{leecorrrelation2022}. 
Here, we evaluate retrieval performance when applying SuperGlobal reranking on top of CVNet-Global. Please note that the other modules introduced in SuperGlobal (e.g. GeM+, Scale-GeM, Regional-GeM) are not included in this section of experiments. As shown in Table \ref{tab:cvnet_supp}, we report that applying SuperGlobal reranking module to CVNet-Global significantly improve the performance in both $\mathcal{R}$Oxford and $\mathcal{R}$Paris datasets. When comparing with CVNet reranking (Table \ref{tab:sota_supp}), using SuperGlobal reranking still shows superior performance in the $\mathcal{R}$Paris dataset.